# EnzChemRED, a rich enzyme chemistry relation extraction dataset

Po-Ting Lai[1,†], Elisabeth Coudert[2,†], Lucila Aimo[2], Kristian Axelsen[2], Lionel Breuza[2], Edouard de Castro[2], Marc Feuermann[2], Anne Morgat[2], Lucille Pourcel[2], Ivo Pedruzzi[2], Sylvain Poux[2], Nicole Redaschi[2], Catherine Rivoire[2], Anastasia Sveshnikova[2], Chih-Hsuan Wei[1], Robert Leaman[1], Ling Luo[3], Zhiyong Lu[1,*], and Alan Bridge[2,*]

[1]National Center for Biotechnology Information (NCBI), National Library of Medicine (NLM), National Institutes of Health (NIH), Bethesda, MD 20894, USA; [2]Swiss-Prot group, SIB Swiss Institute of Bioinformatics, Centre Medical Universitaire, CH-1211 Geneva 4; [3]School of Computer Science and Technology, Dalian University of Technology, 116024, Dalian, China

* Corresponding authors: Alan Bridge (alan.bridge@sib.swiss) and Zhiyong Lu (zhiyong.lu@nih.gov).
† These authors contributed equally to this work.

## Abstract

Expert curation is essential to capture knowledge of enzyme functions from the scientific literature in FAIR open knowledgebases but cannot keep pace with the rate of new discoveries and new publications. In this work we present EnzChemRED, for Enzyme Chemistry Relation Extraction Dataset, a new training and benchmarking dataset to support the development of Natural Language Processing (NLP) methods such as (large) language models that can assist enzyme curation. EnzChemRED consists of 1,210 expert curated PubMed abstracts in which enzymes and the chemical reactions they catalyze are annotated using identifiers from the UniProt Knowledgebase (UniProtKB) and the ontology of Chemical Entities of Biological Interest (ChEBI). We show that fine-tuning pre-trained language models with EnzChemRED can significantly boost their ability to identify mentions of proteins and chemicals in text (Named Entity Recognition, or NER) and to extract the chemical conversions in which they participate (Relation Extraction, or RE), with average $F_1$ score of 86.30% for NER, 86.66% for RE for chemical conversion pairs, and 83.79% for RE for chemical conversion pairs and linked enzymes. We combine the best performing methods after fine-tuning using EnzChemRED to create an end-to-end pipeline for knowledge extraction from text and apply this to abstracts at PubMed scale to create a draft map of enzyme functions in literature to guide curation efforts in UniProtKB and the reaction knowledgebase Rhea.

**Availability:** The EnzChemRED corpus is freely available at https://ftp.expasy.org/databases/rhea/nlp/.

**Contact:** alan.bridge@sib.swiss , luzh@ncbi.nlm.nih.gov

# 1. Introduction

Knowledge of enzyme functions is critical to our understanding of how biological systems function and interact in complex communities (1-3), how genetic variation and disease affect those systems (4-6), and for efforts to engineer those systems to synthesize beneficial compounds such as drugs and biofuels or break down harmful environmental pollutants (7-13). Expert curated knowledgebases such as the UniProt Knowledgebase (UniProtKB) (14,15), Rhea (16), MetaCyc (17), KEGG (18), BRENDA (19), SABIO-RK (20), Reactome (21) and the Gene Ontology (GO) (22) capture knowledge of enzymes and the reactions they catalyze from peer reviewed publications using human- and machine-readable chemical ontologies and cheminformatics descriptors in forms that are Findable, Accessible, Interoperable, and Reusable (FAIR) (23). These databases play a critical role in biological and biomedical research but face significant challenges in keeping pace with the discovery and publication of new enzymes and reactions, with the result that much of our knowledge of how enzymes function remains "locked" in peer-reviewed publications in forms that are difficult for both humans and machines to access.

Natural Language Processing (NLP) methods offer a potential means to accelerate the expert curation of enzyme functions, with large language models based on the transformer architecture such as BERT (24) among the most promising methods developed to date. The models are pre-trained using self-supervised approaches on large corpora of unannotated text using a language modeling objective and can be fine-tuned to perform specific tasks using curated domain-specific corpora in a process of transfer learning (25-27). At the current time there is no freely available curated domain-specific corpus for fine-tuning language models to extract enzyme functions from text, and rule-based (28) and weakly supervised machine learning approaches (29) have been used instead.

In this work we aim to develop a corpus that could be used to fine-tune language models and other NLP methods to assist curators in extracting knowledge of enzymes and their reactions from text. This corpus, EnzChemRED, for <u>Enz</u>yme <u>Chem</u>istry <u>R</u>elation <u>E</u>xtraction <u>D</u>ataset, consists of 1,210 abstracts from PubMed in which the chemical conversions and the enzymes that catalyze them are curated using stable unique identifiers from UniProtKB and the chemical ontology ChEBI (30). We propose a methodology to extract knowledge of enzyme functions from the literature by framing the problem as a series of NLP tasks, beginning with named entity recognition (NER), to identify text spans that denote enzymes and the chemicals they act on, named entity normalization (NEN), to link text spans to database identifiers, and relation extraction (RE), to link mentions of pairs of chemicals that define conversions (binary relations) and to link those conversions to mentions of enzymes that catalyze them (ternary relations). To establish a baseline for future research we present the results from fine-tuning pre-trained language models using EnzChemRED, achieving an $F_1$ score (harmonic mean of precision and recall) of 86.30% for NER, and 86.66% for binary RE and 83.79% for ternary RE. We combined these methods in a prototype end-to-end pipeline that performs literature triage, NER, NEN, and RE, and applied this pipeline to PubMed abstracts to create a draft map of enzyme functions in literature to guide and assist curation efforts in UniProtKB/Swiss-Prot and Rhea. We hope that this benchmark dataset will be of broad utility for NLP researchers and the wider community of knowledgebase developers and biocurators of other resources, and welcome feedback and suggestions for further improvements to EnzChemRED and the methods described here.

## 2. Related Works

Several prior works have described datasets to train and benchmark methods to extract information about chemicals and their relations from scientific literature. Table 1 provides a chronology and summary of their main characteristics. They include datasets that address the problem of chemical NER, such as that of Corbett *et al*. (31), the SCAI dataset of Fluck and colleagues (32), and the CHEMDNER dataset (33), datasets that address chemical NEN, such as the BC5CDR dataset (34) and the Biomedical Relation Extraction Dataset (BioRED) (35), which both map chemical mentions to identifiers from the Medical Subject Headings (MeSH), and the EBED dataset (36), which maps chemical mentions to ChEBI, and datasets that address chemical RE, including the BC5CDR and EBED datasets that link chemicals to chemically induced diseases, the ADE dataset (37), that links drugs and adverse drug effects, the ChemProt (38) and DrugProt (39) datasets, that link chemicals to proteins, and the *n*-ary dataset of Titinsky and colleagues that links drugs, genes, and mutations (40). Of the very few datasets that include chemical-chemical associations, most focus on interactions between drugs, such as the drug combination *n*-ary dataset (40) and the drug-drug interaction (DDI) dataset (41). Only the BioRED and ChEMU lab 2020 dataset (42) include chemical conversions, but the BioRED dataset includes only seventeen chemical conversion pairs, which is too few for any meaningful assessment, while the ChEMU lab 2020 dataset focuses on organic chemistry from patents relevant to drug synthesis, and not on descriptions of enzyme functions. EnzChemRED differs from these prior works in that it focuses specifically on the problem of extracting chemical conversions catalyzed by enzymes, considers not only chemical conversions (binary relations that link pairs of chemicals) but also the enzymes that catalyze them (ternary relations that link conversions to enzymes), and addresses all the steps needed to extract those relations, namely NER, NEN, and RE.

## Table 1. Overview of datasets for chemical NLP

| Year | Dataset | Documents | Chemical, protein, and gene mentions | Unique IDs | Relations |
|---|---|---|---|---|---|
| 2008 | Corbett (31) | 500 abstracts, 42 papers | 11,571 | - | - |
| 2008 | SCAI (32) | 100 abstracts | 1,206 | - | - |
| 2012 | ADE (37) | 300 case reports | 5,063 drugs | - | 6,821 drug adverse effects<br>279 drug dosage |
| 2013 | DDI (41) | 1,025, including texts from DrugBank and abstracts | 18,502 drugs | - | 5,028 drug-drug interactions |
| 2015 | CHEMDNER (33) | 10,000 abstracts | 84,355 chemicals | - | - |
| 2016 | BC5CDR (34) | 1,500 articles | 15,935 chemicals<br>12,850 diseases | 4,409 MeSH | chemically induced diseases |
| 2017 | N-ary drug- gene- mutation (40) | - | - | - | 137,469 drug–gene<br>3,192 drug–mutation |
| 2017 | ChemProt (38) | 2,482 abstracts | 32,514 chemicals<br>30,922 genes | - | chemical-protein |
| 2019 | DrugProt (39) | 5,000 abstracts | 65,561 chemicals<br>61,775 proteins | - | 24,526 chemical-protein |
| 2020 | EBED (36) | 4,200, including abstracts, paragraphs, figure legends, and patents | 16,715 chemicals<br>56,059 genes | 5,161 ChEBI<br>12,563 Entrez | chemically induced diseases |
| 2021 | ChEMU 2020 (42) | 1,500 patent extracts | 17,834 chemicals | - | chemical reaction steps |
| 2022 | BioRED (35,43) | 1,000 abstracts | 4,429 chemicals<br>6,697 genes | 651 MeSH<br>1643 Entrez | chemical-(chemical/disease gene/variant) |
| 2024 | EnzChemRED (this work) | 1,210 abstracts | 18,887 chemicals<br>13,028 proteins | 3,155 ChEBI<br>2,569 UniProtKB | chemical-chemical ([chemical-chemical]-protein) |

# 3. Materials and Methods

Section 3.1 describes the development of the EnzChemRED dataset, which is the focus of this paper. Sections 3.2 to 3.5 describe the development of a prototype end-to-end NLP pipeline for enzyme functions that makes use of EnzChemRED for fine-tuning and benchmarking. The four main steps of this end-to-end NLP pipeline are:

1. Literature triage, to identify relevant papers about enzyme functions (Section 3.2).
2. NER, to tag chemical and protein mentions (Section 3.3).
3. NEN, to link chemical and protein mentions to stable unique database identifiers (Section 3.4).
4. RE, to extract information about enzymes and the chemical conversions that they catalyze (Section 3.5).

Section 3.6 describes the combination of methods to create the end-to-end pipeline, as well as methods to process and visualize the output.

## 3.1 Development of the EnzChemRED corpus

### 3.1.1 Selection of abstracts for curation in EnzChemRED

To build EnzChemRED we selected papers curated in UniProtKB/Swiss-Prot that describe enzyme functions. We queried the UniProt SPARQL endpoint (https://sparql.uniprot.org/) to identify papers that provided experimental evidence used to link UniProtKB/Swiss-Prot protein sequence records to reactions from Rhea that involve only small molecules (excluding papers linked to Rhea reactions that involve proteins and other macromolecules) (Figure 1). UniProtKB/Swiss-Prot uses evidence tags and the Evidence and Conclusions Ontology (ECO) (44) to denote provenance and evidence for functional annotations; our SPARQL query selected papers linked to Rhea annotations in UniProtKB/Swiss-Prot with evidence tags with experimental evidence codes from ECO – such as "ECO:0000269", which denotes "experimental evidence used in manual assertion". We also narrowed the selection of abstracts to those including mentions of at least one pair of reactants found in Rhea, and to those having a score of at least 0.9 according to our LitSuggest (45) model for abstracts relevant to enzyme function (see Section 3.2 of Materials and Methods). This LitSuggest score threshold is exceeded by 99% of abstracts of papers curated in Rhea. We selected 1,210 abstracts and divided these into 11 groups of 110 abstracts for curation by our team of expert curators.

### 3.1.2 Curation of chemical and protein mentions in EnzChemRED

Curation of abstracts in EnzChemRED was performed using the collaborative curation tool TeamTat (www.teamtat.org) (46) following a protocol based on that developed for the curation of the BioRED dataset (available at https://ftp.ncbi.nlm.nih.gov/pub/lu/BioRED/), with modifications described below. We curated five types of entity in EnzChemRED: Chemical, Protein, Domain, MutantEnzyme, and Coreference, which are described below, with examples shown in Figure 2.

- **Chemical**: a mention of a chemical entity – including chemical structures, chemical classes, and chemical groups. Where possible we normalized chemical mentions to identifiers from ChEBI, which provides chemical structure information, and which is used to describe chemical entities in both Rhea and UniProtKB. Where no ChEBI identifier was available we used MeSH. A small number of chemical mentions have no mapping to either resource.
- **Protein**: a mention of a protein, or family of proteins, normalized to UniProtKB accession numbers (UniProtKB ACs). As in BioRED, we included gene names in our annotation, which we also normalized to UniProtKB ACs.
- **Domain**: a mention of a protein domain, normalized to the UniProtKB ACs of the protein in which the domain occurs.
- **MutantEnzyme:** a mention of a mutant protein, normalized to the UniProtKB accession number of the wild-type protein.
- **Coreference:** a reference to a protein or chemical that is explicitly named elsewhere in the abstract (authors may name a protein in one sentence and describe its function in another), and normalized to the identifier used for the mention referenced.

We curated all Chemical and Protein mentions found in abstracts, irrespective of whether those mentions were part of descriptions of enzymatic reactions or not. We did not systematically curate Domain, MutantEnzyme and Co-reference mentions, but focused on those that participate in enzymatic reactions. For this reason, we did not consider these three types of mentions in our evaluations of NER, NEN, and RE performance, but these could serve as valuable annotations for more extensive dataset development in the future.

### 3.1.3 Curation of chemical conversions in EnzChemRED

We based our schema for the curation of relations relevant for enzyme functions in EnzChemRED on that developed in BioRED, but with three major alterations.

First, we defined two additional relation types in EnzChemRED. BioRED captures chemical reactions by using the relation "Conversion" to link pairs of reactants. In EnzChemRED we also added "Indirect_conversion" and "Non_conversion", giving three possible relations (Table 2) that are defined as follows:

- **Conversion**: links two chemicals that, according to the text, may participate on opposite sides of a reaction equation – such as one substrate and one product.
- **Indirect_conversion**: links two chemicals that, according to the text, can interconvert, but not directly – such as conversions involving the first and last chemical in a pathway. While these kinds of relations will not give rise to enzyme function annotations, they constitute a significant but minor fraction of relations in the EnzChemRED dataset (see Section 4.1).
- **Non_conversion**: links two chemicals that, according to the text, were experimentally tested but did not interconvert at all (at least under the experimental conditions used). These are the rarest type of relation in EnzChemRED.

Second, while BioRED features only binary pairs, in EnzChemRED we also introduced ternary tuples, which allow us to link mentions of enzymes to the "Conversions" they catalyze. We assign each enzyme the role of "Converter".

Third, we modified the granularity of annotations in BioRED. While BioRED provides document-level relation pairs, EnzChemRED provides relations annotated at the level of individual mentions and sentences.

**Table 2. Examples of curated relations in EnzChemRED.** Chemical mentions and protein mentions are denoted by the numbered subscripts "c" and "p" respectively.

| Relation type | Example | Explanation |
| --- | --- | --- |
| Conversion | "**D-Dopachrome tautomerase** $_{P1}$ converts **2-carboxy-2,3-dihydroindole-5,6-quinone** $_{c1}$ (**D-dopachrome**) $_{c2}$ into **5,6-dihydroxyindole** $_{c3}$." – PMID: 9480844 | P1 can convert C1 to C3, and C2 is a synonym for C1. We therefore curate two "Conversion" relations of (C1, C3) by P1 as "Converter", and of (C2, C3) by P1 as "Converter". |
| Indirect_conversion | "Cell extracts of Brucella abortus (British 19) catabolized **erythritol** $_{c1}$ through a series of phosphorylated intermediates to **dihydroxyacetonephosphate** $_{c2}$ and **CO-2** $_{c3}$." – PMID:163226 | C1 can be converted to C2 and C3, but indirectly, via a series of intermediates; no enzyme is mentioned. We therefore curate both (C1, C2) and (C1, C3) as "Indirect_conversion", with no "Converter". |
| Non_conversion | *"In the amination direction, they catalyze the conversion of* **mesaconate** $_{c1}$ *to yield only* **(2S,3S)-3-methylaspartic acid** $_{c2}$, *with no detectable formation of* **(2S,3R)-3-methylaspartic acid** $_{c3}$." – PMID:19670200 | C1 can be converted to C2, but not C3. We therefore curate (C1, C2) as a "Conversion", and (C1, C3) as a "Non_conversion". |

### *3.1.4 Curation of the EnzChemRED corpus – workflow*

Figure 3 outlines the curation workflow for EnzChemRED; we describe the main steps below.

**Pre-Annotation of chemicals and proteins.** We used PubTator (47,48) to pre-tag chemical and gene/protein mentions in the 1,210 abstracts of EnzChemRED prior to their curation. PubTator assigns MeSH IDs for chemical mentions and Entrez IDs for gene and protein mentions, which we converted to ChEBI IDs and UniProtKB ACs using MeSH-to-ChEBI and Entrez-to-UniProtKB mapping tables. ChEBI provides multiple distinct identifiers for different protonation states of a given chemical compound, so we mapped all ChEBI IDs to those of the major protonation state at pH 7.3 – the form used in UniProtKB and Rhea – using a mapping file created for this purpose by Rhea (the file "chebi_pH7_3_mapping.tsv", which is available at https://www.rhea-db.org/help/download).

**Curation by human experts.** Curation was performed using TeamTat (Figure 4) by a team of 11 professional curators with expertise in biochemistry and the curation of UniProtKB/Swiss-Prot and Rhea. Curators were required to review all PubTator tagging results for gene/protein and chemical mentions (both text spans and IDs), correct them as necessary, and add missing protein and chemical annotations and identifiers. Curators were allowed to use external information sources, including the full text of the article, as well as knowledge resources such as UniProtKB, Rhea, ChEBI, MeSH, and PubChem (49), when curating chemical and protein mentions. Following curation of all protein and chemical mentions, curators were then required to link chemical mentions that participate in relations of the type "Conversion", "Indirect_conversion", or "Non_conversion", thereby creating binary (chemical-chemical) pairs, as well as mentions of enzymes that catalyze those conversions ("Converter") if applicable, creating ternary [enzyme-(chemical-chemical)] tuples. Curators were prohibited from using external information, such as the full text of the publication or prior knowledge of the chemistry or enzymes involved in the reactions, when annotating relations of any type, or linking converters to conversions. Put another way, all the evidence needed to create a conversion, and to link a converter to it, had to be contained in the abstract itself, either within one sentence, or across multiple sentences.

We divided the curation process into four phases.

1. **Phase 1.** We provided each curator with 10 abstracts for familiarization with the curation workflow and guidelines. Following curation, the abstracts were frozen, and a copy was made, which was reviewed and corrected by a second curator. The curation team then met to discuss curation issues, revise guidelines, and finalize the set of abstracts from phase 1. The output from phase 1 consisted of 110 curated abstracts, each reviewed and where necessary revised, and a set of revised guidelines.
2. **Phase 2.** We provided each curator with 50 additional abstracts for curation. Curated abstracts were frozen, and a copy was made, which was reviewed and corrected by a second curator. The curation team then met to discuss curation issues, revise guidelines, and finalize the set of abstracts from phase 2. The output from phase 2 consisted of a further 550 curated abstracts, each reviewed and where necessary revised, and a set of revised guidelines.
3. **Phase 3.** We provided each curator with 50 additional abstracts for curation. These were not reviewed after curation. The output from phase 3 consisted of a further 550 curated abstracts that had not been reviewed.
4. **Phase 4.** We performed a round of "model guided" re-curation of abstracts, using a "preliminary" BioREx model (Section 3.5) to identify potential curation errors, such as missed chemical conversions. We trained this model using the set of 1,210 abstracts curated in phases 1-3 and used it to perform RE on the entire dataset of 1,210 abstracts. We identified potential false positive (FP) or false negative (FN) predictions in 575 of the 1,210 abstracts. Each potential FP or FN prediction identified by the preliminary BioREx model was then examined by two curators, who were free to compare and discuss their interpretations of the models' predictions. In some cases, the potential FP and FN errors from the model were deemed to be correct and were re-curated as TP or TN as appropriate, and the curation guidelines were updated if needed. We used this Phase 4 EnzChemRED dataset to train our final models for NER (see Section 3.4) and RE (see Section 3.5).

### 3.2. Literature triage

We used LitSuggest (https://www.ncbi.nlm.nih.gov/research/litsuggest/) (45), a web-based machine-learning framework for literature recommendations, to triage papers relevant for enzyme functions in our end-to-end pipeline, and as an additional check on relevance when selecting abstracts for EnzChemRED curation (see Section 3.1 of Materials and Methods).

Positive training examples for LitSuggest consisted of abstracts from papers that provided experimental evidence used to annotate enzymes in UniProtKB/Swiss-Prot with Rhea reactions (dataset created November 7th, 2020). We defined papers that provided experimental evidence as those linked to an evidence tag with an experimental evidence code from the Evidence and Conclusions Ontology (ECO) (44), such as "ECO:0000269", which denotes "experimental evidence used in manual assertion". We excluded abstracts of papers linked to Rhea reactions that involve proteins and other macromolecules, such as DNA. This exclusion criterion strongly reduced the propensity of LitSuggest models to retrieve papers about signaling pathways, where proteins are modified by enzymes.

We trained and tested LitSuggest models using a set of 9,055 positive abstracts split into 5 sets of 7,244 positive abstracts for training and 1,811 positive abstracts for testing, with 14,488 negative abstracts for training selected at random from PubMed using the LitSuggest curation interface. LitSuggest models provide a score of 0-1 for each abstract, with scores above 0.5 denoting that the abstract is relevant (belongs to the same class as the positive training data). The five LitSuggest models had a mean recall of 98% when tasked with classifying the 1,811 abstracts left out. To test precision and recall using more realistic ratios of relevant and irrelevant literature, we also performed "spike-in" tests that mixed 250 relevant papers (from the set of 1,811 abstracts left out) with a set of 100,000 abstracts selected at random from PubMed using NCBI e-utils (so a ratio of 400:1 irrelevant to relevant abstracts). At a score threshold of 0.8, our best performing LitSuggest model had precision of 90.1%, sensitivity of 94.8%, and $F_1$ score of 92.4% in these "spike-in" tests. Swiss-Prot curators now use this LitSuggest model as a tool to triage literature for curation of protein sequences in UniProtKB/Swiss-Prot with reactions from Rhea on a weekly basis. It is available at https://www.ncbi.nlm.nih.gov/research/litsuggest/project/5fa57e75bf71b3730469a83b. We also used this model in the first step of our end-to-end pipeline for enzyme function extraction from PubMed abstracts.

### 3.3. Named Entity Recognition (NER)

We tested four different pre-trained language models (PLMs) for NER, namely Bioformer (50), PubMedBERT (27), AIONER-Bioformer (51), and AIONER-PubMedBERT (51). We performed 10-fold cross validation for each model using EnzChemRED, fine-tuning the PLMs using the training set partition and evaluating them on the test set partition. AIONER (https://github.com/ncbi/AIONER) is an all-in-one scheme-based biomedical NER tool that integrates multiple resources into a single task via adding task-oriented tagging labels. AIONER utilizes a simple but effective way to combine diverse entity-type datasets into one, allowing the model to learn different synonyms present in diverse texts. AIONER performs optimally on 14 BioNER benchmark datasets, such as BioRED, BC5CDR (52), GNormPlus/GNorm2 (53), NLM-Gene (54), and NLM-Chem (55). It has been shown to achieve competitive performance with Multi-Task Learning (MTL) methods for multiple NER tasks, while being more stable and requiring fewer model modifications.

To fine tune AIONER for our application we used the source code implementation of AIONER. To begin with, we transformed the EnzChemRED corpus to align with AIONER's input representation in the IOB2 tagging scheme. This format is like the standard IOB2, but it includes additional task boundary tags at the start and end of each sentence. We illustrate the training process for the AIONER model using EnzChemRED in Figure 5. The initial step involves the creation of two unique tokens, <Reaction> and

</Reaction>, which indicate the beginning and end of each sentence. Subsequently, we employ spaCy (https://spacy.io/) for sentence detection and tokenization. This allows us to convert our entire biochemical reaction text corpus into the IOB2 tagging scheme. AIONER offers a unique approach to label formulation by further categorizing "O" labels, which in the IOB2 tagging scheme represent "others." AIONER divides "O" into more detailed labels, each corresponding to a particular task, such as "O-All" for the BioRED task and "O-Gene" for the GNormPlus and NLM-Gene tasks. In our case, we use "O-Reaction" to signify the NER task, giving five labels for our dataset, namely "O-Reaction," "B-Gene," "I-Gene," "B-Chemical," and "I-Chemical". The "B-" and "I-" labels in the AIONER scheme are the same as the traditional BIONER scheme and denote "begin" and "inside" tokens. Once the dataset is converted to the IOB2 tagging scheme, the pre-trained AIONER model can be downloaded. Our model is optimized using the fine-tuning script provided by AIONER on GitHub (see above). A similar procedure is followed during the testing phase.

### 3.4. Named Entity Normalization (NEN)

We used the Multiple Terminology Candidate Resolution (MTCR) pipeline to map chemical mentions in abstracts to ChEBI and MeSH IDs (NEN). MTCR is a structured approach for linking entities in the biomedical domain, including chemical terminologies. A similar process – referred to as sieve-based entity linking – has been described by D'Souza and Ng (56). There are three main steps in the MTCR pipeline: abbreviation resolution, candidate lookup, and post-processing.

1. During the abbreviation resolution phase, the pipeline identifies pairs of short and long forms in each document using the Ab3P Abbreviation Resolution tool (57). Short forms are then expanded into long forms before looking up (so "TPP" to "triphenyl phosphate").
2. The candidate lookup starts with a precise lookup, and proceeds to higher recall queries, stopping once a match is found. We lower the case of the text and strip non-alphanumeric characters from it. The procedure for this phase involves searching for exact texts, processed texts, and stemmed texts in the target lexicon, if necessary, as well as in all lexicons.
3. The results are then mapped to the target lexicon. Mappings can take place in two ways: 'single', where terms are mapped directly to the target, and 'pivot', where terms are mapped directly to the target through 'pivot' identifiers shared with another lexicon, such as the International Chemical Identifier (InChI) (or it's hashed form, the InChIKey) (https://www.inchi-trust.org) (58), the SMILES (Simplified Molecular-Input Line-Entry System) (http://opensmiles.org), a linear notation for chemical structures, the Chemical Abstracts Service (CAS) number, or others.

When the target terminology is broad, such as MeSH, the post-processing phase includes removing annotations for mentions identified as non-chemical. In addition, it resolves ambiguous mentions with unambiguous ones. The MTCR pipeline has been benchmarked for the BioCreative VII NLM Chem task (59) with MeSH as the target terminology. BlueBERT (60) showed higher NER performance, but MTCR demonstrated outstanding NEN performance, with a precision of 81.5% and a recall of 76.4%, with only 29% of teams outperforming MTCR in NEN.

## 3.5. Relation extraction (RE)

We frame the problem of extracting pairs of chemical reactants as one of relation classification.

- For each binary (chemical-chemical) pair in each sentence, the goal is to predict the type of relation between the pair of chemical mentions.
- For each ternary [enzyme-(chemical-chemical)] tuple in each sentence, the goal is to predict the type of relation between the pair of chemical mentions, and to correctly identify the enzyme (Converter).

Valid relation types for binary pairs and ternary tuples are "Conversion", "Indirect_conversion", and "Non_conversion", which are curated (see Section 3.1.3), and "None", which is assigned automatically during evaluation. For binary pairs, "None" is assigned to all pairs of chemical mentions that are not curated using one of the three valid relation types. For ternary tuples, "None" is assigned to all ternary tuples that include pairs of chemical mentions not curated using one of the three valid relation types, and to all ternary tuples that include pairs of chemicals curated with a valid relation but that also include a protein mention that was not linked to them by a curator (i.e. it is not the enzyme responsible).

We performed relation classification using PubMedBERT and BioREx (61), which is a PubMedBERT model trained on the BioRED dataset and eight other common biomedical RE benchmark datasets. BioREx offers a reliable and effective approach to chemical reaction extraction and has shown consistently high performance for relation classification across seven different entity pairs. The output from the PubMedBERT and BioREx models is a vector that corresponds to our four relation types, which is derived from the [CLS] vector, and which is fed in turn into a SoftMax function, so that each binary pair or tuple is assigned a score for each of the four relation types. We performed 10-fold cross validation for each model using EnzChemRED, fine-tuning the models using the training set partition and evaluating them on the test set partition.

Figure 6 illustrates the process of fine-tuning on EnzChemRED using BioREx. EnzChemRED is annotated at both mention and sentence levels, with locations specified, unlike BioRED, which uses document-level annotation, and where relations are given in the format of ID pairs without specifying the exact locations of the entity mentions involved. We therefore adjust the fine-tuning procedure used in BioRED, replacing the [Corpus] tag with a [Reaction] tag and using individual sentences as input rather than full documents.

Figure 7 illustrates an example input representation of a ternary tuple. The classification of ternary tuples follows similar rules to that for binary pairs. We insert additional boundary tags, "<P>" and "</P>", to denote the enzyme in the input instance, but otherwise follow the same procedure as for binary pair RE. As with binary pairs, ternary tuples are annotated at both sentence and mention levels, such that if the same enzyme appears more than once in a sentence, each occurrence would be treated as a different instance.

## 3.6. End-to-end pipeline

We combined the best performing methods for NER (AIONER-PubMedBERT, fine-tuned using EnzChemRED) and RE (BioREx, fine-tuned using EnzChemRED) with MTCR for NEN to create a prototype end-to-end pipeline for enzyme function extraction from text. We applied this pipeline to EnzChemRED abstracts for cross validation purposes, and to relevant PubMed abstracts (up to December 2023) identified using the LitSuggest model described in Section 3.2 to map enzyme functions in literature. The latter necessitated comparison of chemical pairs extracted from PubMed abstracts to pairs of chemical reactants from Rhea, which was accomplished as follows. To create the set of Rhea pairs for comparison, we extracted pairs of chemical reactants from Rhea using a heuristic procedure in which we removed the top 100 most frequently occurring compounds in Rhea reactions such as water, oxygen, and protons, and then enumerated all possible pairs of the remaining compounds for each Rhea reaction. We also removed pairs of identical ChEBI IDs from the Rhea set, which in Rhea can occur as part of transport reactions. To prepare the chemical pairs extracted from PubMed abstracts for comparison to Rhea, we first normalized their ChEBI IDs to those representing the major microspecies at pH7.3, the form used in Rhea, removed pairs that include any of the top 100 most frequently occurring compounds in Rhea reactions, and removed pairs where both members had the same ChEBI ID. This can occur due to errors in NER and NEN, where erroneous text spans can cause distinct but related chemical names to be mapped to the same identifier. After processing we compared the degree of overlap in the two sets (chemical pairs from Rhea reactions and PubMed abstracts) using their ChEBI IDs.

### 3.7. Visualization of chemical pairs from PubMed abstracts and Rhea

To visualize chemical pairs from PubMed abstracts and Rhea we created Tree Maps (TMAPs) using code from http://tmap.gdb.tools as described (62), and clustered chemical pairs in TMAPs according to their Differential Reaction Fingerprint (DRFP), calculated according to the method of Probst (63).

We used the degree of atom conservation between the members of each chemical pair to filter the output of our end-to-end NLP pipeline. To calculate atom conservation, we first convert molecular structures into graphs by replacing all bond types with single bonds. This ensures a standardized representation of molecular structures, simplifying subsequent analyses. We then compute the Maximum Common Substructure (MCS) using the rdkit.Chem.rdFMCS.FindMCS function (from the open-source cheminformatics toolkit RDKit, at www.rdkit.org) with a permissive ring fusion parameter. The MCS represents the largest common atomic framework shared by the two molecules (after conversion into a graph of atoms linked by single bonds). The atom conservation is the average of the percentage of common atoms, as given by:

$$\% \text{ atom conservation} = \tfrac{1}{2} \times (n_{mcs}/n_l + n_{mcs}/n_r) \times 100$$

where $n_{mcs}$ is the number of atoms in the maximum common substructure, $n_l$ is the number of atoms in the molecule on the left side of the pair, and $n_r$ is the number of atoms in the molecule on the right side of the pair. This metric provides a standardized measure of structural similarity, facilitating the comparison of chemical compounds in each pair.

## 4. Results

## 4.1 Dataset statistics and inter-annotator agreement (IAA)

Table 3 provides an overview of the EnzChemRED dataset, highlighting key statistics including counts of documents, entity mentions, and binary and ternary relations. The dataset comprises 1,210 abstracts, with 31,915 curated entity mentions and 5,724 unique protein and chemical IDs. Mentions of chemical entities, with 18,887 mentions and 3,155 unique ChEBI IDs, are more common than those of proteins, with 13,028 entities and 2,569 unique UniProtKB ACs, while the most common type of relation linking pairs of chemicals is the "Conversion". The inter-annotator agreement (IAA) of our dataset stands at 92.82% for NER + NEN, indicating strong agreement among annotators when selecting text spans and identifiers, and at 87.03% for binary pair annotation, which requires that both spans and identifiers match.

**Table 3. EnzChemRED summary statistics.**
The EnzChemRED corpus consists of 1,210 curated abstracts.

| Type | | Entity mentions, pairs, tuples | Unique IDs, pairs, or tuples |
|---|---|---|---|
| Entity | All | 31,915 | 5,724 |
| | Chemical | 18,887 | 3,155 |
| | Protein | 13,028 | 2,569 |
| Binary pair | All | 4,817 | 2,679 |
| | Conversion | 4,386 | 2,434 |
| | Indirect_conversion | 411 | 292 |
| | Non_conversion | 20 | 17 |
| Ternary tuple | All | 4,195 | 2,120 |
| | Conversion | 3,966 | 1,997 |
| | Indirect_conversion | 229 | 133 |
| | Non_conversion | 0 | 0 |

## 4.2 Evaluation of NER methods using EnzChemRED

We evaluated the effects of fine-tuning using EnzChemRED on NER using four pre-trained language models: Bioformer, PubMedBERT, AIONER-Bioformer, and AIONER-PubMedBERT. Bioformer and PubMedBERT are the base models used to test the effect of fine-tuning using EnzChemRED alone; AIONER-Bioformer and AIONER-PubMedBERT were pretrained on a wide range of chemical and gene NER datasets. For evaluation, we used the $F_1$ score and considered the PMID, entity type (chemical or protein), and the start and end characters of named entities. An entity detected by the NER method is considered a true positive only if both the entity type and the start and end character positions match.

Table 4 shows NER performance for each of the four models. All models performed well in chemical NER, with similar performance for models that were either trained with AIONER, or fine-tuned using EnzChemRED, suggesting that the chemical datasets included in AIONER and the annotations in EnzChemRED provide similar benefits to the models for NER. All models performed less well in protein NER, which may be due to the greater variation in text spans for genes and proteins, but fine-tuning with EnzChemRED significantly improved the performance of all four models for proteins too. The best performing model overall is AIONER-PubMedBERT fine-tuned using EnzChemRED, with $F_1$ scores of 87.26% for chemical NER and 84.93% for gene/protein NER, which is comparable to SOAT performance for chemical NER ($F_1$ score 84.79%) (64) on the NLM-Chem dataset (59) and for gene/protein NER ($F_1$ score 86.70%) (47) on the GNormPlus dataset (65).

**Table 4. NER performance of pre-trained language models on EnzChemRED.**
P, precision; R, recall; F, $F_1$ score.

| PLM | Chemical | | | Protein | | | Overall | | |
| --- | --- | --- | --- | --- | --- | --- | --- | --- | --- |
| | P | R | F | P | R | F | P | R | F |
| AIONER-Bioformer | 86.13 | 85.80 | 85.97 | 76.37 | 78.79 | 77.56 | 82.08 | 82.95 | 82.51 |
| AIONER-PubMedBERT | 87.10 | 85.52 | 86.30 | 79.53 | 74.99 | 77.19 | 84.10 | 81.23 | 82.64 |
| Bioformer +EnzChemRED | 85.83 | 86.87 | 86.35 | 83.13 | 84.75 | 83.93 | 84.73 | 86.01 | 85.36 |
| PubMedBERT +EnzChemRED | 86.92 | 87.21 | 87.07 | 83.19 | 85.07 | 84.11 | 85.38 | 86.33 | 85.86 |
| AIONER-Bioformer +EnzChemRED | 86.20 | 86.95 | 86.58 | 81.90 | 85.35 | 83.59 | 84.41 | 86.30 | 85.35 |
| AIONER-PubMedBERT +EnzChemRED | 87.16 | 87.37 | **87.26** | 83.80 | 86.09 | **84.93** | 85.77 | 86.85 | **86.30** |

### 4.3 Evaluation of RE methods using EnzChemRED

We evaluated the effects of fine-tuning using EnzChemRED on RE using two pre-trained language models: PubMedBERT and BioREx. To evaluate binary pair RE using EnzChemRED we consider the PMID, the start and end character positions of the chemical pair within the sentence, their ChEBI or MeSH IDs, and the relation type. For ternary tuple RE using EnzChemRED we also consider the start and end character positions of the protein mentions within the sentence and the UniProtKB AC. For both binary pair and ternary tuple evaluation, we used two types of classification: binary and multi-class classification. Binary classification considers "Conversion" and "Indirect_conversion" as equivalent, while multi-class classification considers them as distinct. For training purposes, a "None" relation type is utilized, which is assigned to all chemical pairs that are not curated, and which are presumed true negatives. Chemical and protein mentions that lack identifiers are also considered in the evaluations, with their IDs being treated as empty strings.

Table 5 shows RE performance (both binary pairs and ternary tuples). Both baseline models PubMedBERT and BioREx were poor predictors of binary and multi-class relations for binary and ternary tuples, but the performance of both models increased significantly after fine-tuning with EnzChemRED, with BioREx achieving an $F_1$ score of 86.66% for binary relation classification for chemical pairs, and consistently outperforming PubMedBERT for all RE tasks. We therefore chose to employ BioREx in our end-to-end pipeline. Performance of both models decreased slightly as the complexity of the classification task increased, with reduced $F_1$ score for multi-class relation classification, which requires identification of conversions that require multiple steps in pathways, and for ternary tuples relative to binary pairs, which require that the enzyme be correctly identified.

**Table 5. RE performance of pre-trained language models on EnzChemRED.**
P, precision; R, recall; F, $F_1$ score.

| Model | Classification | Binary pair | | | Ternary tuple | | |
|---|---|---|---|---|---|---|---|
| | | P | R | F | P | R | F |
| PubMedBERT | Binary class | 21.83 | 79.16 | 34.23 | 12.92 | 94.21 | 22.72 |
| | Multi-class | 19.64 | 71.21 | 30.79 | 12.19 | 88.89 | 21.44 |
| BioREx | Binary class | 20.59 | 98.34 | 34.05 | 13.29 | 99.74 | 23.46 |
| | Multi-class | 19.33 | 89.31 | 31.78 | 12.57 | 94.28 | 22.18 |
| PubMedBERT +EnzChemRED | Binary class | 83.18 | 87.79 | 85.43 | 80.16 | 83.03 | 81.57 |
| | Multi-class | 75.73 | 79.93 | 77.77 | 76.55 | 79.28 | 77.89 |
| BioREx +EnzChemRED | Binary class | 85.20 | 88.17 | **86.66** | 83.13 | 84.46 | **83.79** |
| | Multi-class | 77.49 | 80.20 | **78.82** | 79.40 | 80.67 | **80.23** |

### 4.4 Analysis of BioREx RE errors in EnzChemRED

While BioREx performs well in binary pair and ternary tuple relation classification on EnzChemRED, analysis of error cases may identify areas for further improvement. We randomly selected and analyzed a set of around 50 false positive (FP) cases (binary and ternary), where BioREx predicts a relation that was not curated in EnzChemRED, and 50 false negative (FN) cases (binary and ternary), where BioREx failed to predict a relation that was curated in EnzChemRED. Broadly speaking there appear to be two main categories of FP predictions and two main categories of FN predictions, with examples provided below. In each example chemical mentions and protein mentions are denoted by the numbered subscripts "*c*" and "*p*" respectively.

**FP predictions for binary pairs**

The first common category of FP predictions (~20% of all FPs) occurred in sentences with at least two chemical pairs, with all substrates listed first in order, followed by all products in order, and the two lists linked using the term "respectively", as in this example:

> "Wild-type strain L108 and **mdpJ** $_{P1}$ knockout mutants formed **isoamylene** $_{C1}$ and **isoprene** $_{C2}$ from **TAA** $_{C3}$ and **2-methyl-3-buten-2-ol** $_{C4}$, respectively." (PMID: 22194447)

This sentence provides evidence for two curated instances of "Conversion", isoamylene (C1) from TAA (C3), and isoprene (C2) from 2-methyl-3-buten-2-ol (C4). Our BioREx model correctly identified both, but also predicted a third conversion, isoamylene (C1) from 2-methyl-3-buten-2-ol (C4), which constitutes an FP prediction.

The second common category of FP predictions (~23%) occurred in sentences with multiple chemical mentions, and where the chemical mentions involved in the FP predictions also participated in other curated TP "Conversion", as in this example:

> "Most methanogenic Archaea contain an unusual cytoplasmic fumarate reductase which catalyzes the reduction of **fumarate** $_{C1}$ with **coenzyme M** $_{C2}$ (**CoM-S-H** $_{C3}$) and **coenzyme B** $_{C4}$ (**CoB-S-H** $_{C5}$) as electron donors forming **succinate** $_{C6}$ and **CoM-S-S-CoB** $_{C7}$ as products." (PMID: 9578488)

This sentence provides evidence for the conversion of fumarate (C1), coenzyme M (C2) (CoM-S-H (C3)) and coenzyme B (C4) (CoB-S-H (C5)) to succinate (C6) and CoM-S-S-CoB (C7). Our BioREx model identified all these relations but also erroneously identified one "Conversion" involving fumarate (C1) and CoM-S-H (C3). These are in fact a pair of substrates and do not convert one to the other.

The remainder of FP cases did not fall into any single clearly definable category.

**FN predictions for binary pairs**

The most common category of FN cases (~43%) represent instances of missed "Indirect_conversion", as in this example:

> "We provide NMR and crystallographic evidence that the PucG $_{P1}$ protein from Bacillus subtilis catalyzes the transamination between an unstable intermediate ( (S)-ureidoglycine $_{C1}$ ) and the end product of **purine** $_{C2}$ catabolism ( **glyoxylate** $_{C3}$ ) to yield oxalurate $_{C4}$ and glycine $_{C5}$." (PMID: 20852637)

This sentence provides evidence for a curated "Indirect_conversion" between purine (C2), which is the input for the catabolic pathway that yields glyoxylate (C3) as a product, which BioREx failed to identify. BioREx did correctly identify the "Conversion" relations that link participants in the reaction in which (S)-ureidoglycine (C1) and glyoxylate (C3) are converted to oxalurate (C4) and glycine (C5). Note that "Indirect_conversion" relations do not provide direct knowledge of reactions and enzyme functions, and so FN cases will not have a major impact on our goal of identifying enzyme functions in literature.

The second major category of FN cases (~20%) are those where the text describes multiple substrates and products of a reaction, but where our BioREx model fails to classify all the possible pairwise relations between them, as in this example:

> *"The purified enzyme catalyzed transacetylation of the acetyl group not only from **PAF** $_{C1}$ to **lysoplasmalogen** $_{C2}$ forming **plasmalogen analogs of PAF** $_{C3}$, but also to **sphingosine** $_{C4}$ producing **N-acetylsphingosine** $_{C5}$ (**C2-ceramide** $_{C6}$)."* (PMID:10085103)

This sentence describes the conversion of PAF (C1) and lysoplasmalogen (C2) to plasmalogen analogs of PAF (C3), and of PAF (C1) and sphingosine (C4) to acetylsphingosine (C5) (C2-ceramide (C6)). Our BioREx model correctly identified the binary conversions of lysoplasmalogen (C2) to plasmalogen analogs of PAF (C3) but failed to identify the curated binary "conversions" involving PAF (C1).

**FP and FN predictions for ternary tuples**

The classification of ternary tuples by BioREx suffered from many of the types of errors as the classification of binary pairs, as in this example of a FP ternary tuple, where multiple conversions and enzymes are found in the same sentence:

> *"The proteins encoded by **YhfQ** $_{P1}$ and **YhfN** $_{P2}$ were overexpressed in E. coli, purified, and shown to catalyze the ATP $_{C1}$-dependent phosphorylation of fructoselysine $_{C2}$ to a product identified as fructoselysine 6-phosphate $_{C3}$ by 31P NMR ( YhfQ $_{P3}$ ), and the reversible conversion of **fructoselysine 6-phosphate** $_{C4}$ and water to lysine and **glucose 6-phosphate** $_{C5}$ ( **YhfN** $_{P4}$ )."* (PMID:12147680)

The sentence provides evidence for the conversion of fructoselysine 6-phosphate (C4) to glucose 6-phosphate (C5), by YhfN (P4). BioREx erroneously classified the relation linking this pair of chemical mentions to the protein mention YhfQ (P1) as a "Conversion" when the correct enzyme is YhfN (P4).

FN ternary tuples were also observed when a single protein catalyzed multiple conversions:

> *"We conclude that **acs1** $_{P1}$ encodes a bifunctional enzyme that converts **ribulose 5-phosphate** $_{C1}$ into **ribitol 5-phosphate** $_{C2}$ and further into **CDP-ribitol** $_{C3}$, which is the activated precursor form for incorporation of ribitol 5-phosphate into the H. influenzae type a capsular polysaccharide."* (PMID: 10094675)

Our BioREx model correctly identified the conversion of ribulose 5-phosphate (C1) to ribitol 5-phosphate (C2), and of ribitol 5-phosphate (C2) to CDP-ribitol (C3) but failed to link the second of these conversions to acs1 (P1).

### 4.5 Evaluation of end-to-end pipeline on EnzChemRED

We combined the best performing methods for NER (AIONER-PubMedBERT + EnzChemRED) and RE (BioREx + EnzChemRED) with MTCR for NEN to create a prototype end-to-end pipeline for enzyme function extraction from text. This achieved precision of 61.54%, recall of 37.93%, and $F_1$ score of 49.39% for the task of extraction and binary classification of binary chemical pairs from EnzChemRED abstracts (Table 6).

**Table 6. Performance of end-to-end pipeline combining NER, NEN, and RE on EnzChemRED.**
P, precision; R, recall; F, $F_1$ score.

| Task | End-to-end, NER – NEN – RE |
|---|---|

|  | ChEBI matching required | P | R | F |
|---|---|---|---|---|
| Binary pair, binary classification | ChEBI exact | 61.96 | 40.91 | **49.28** |
|  | ChEBI relaxed | 62.61 | 41.17 | **49.55** |
| Ternary tuple, binary classification | ChEBI exact | 25.39 | 17.45 | **20.69** |
|  | ChEBI relaxed | 25.46 | 17.50 | **20.74** |

Significant losses in both precision and recall occurred during the normalization of chemical mentions to identifiers from the ChEBI ontology, with precision of 78.46%, recall of 67.29%, and $F_1$ score of 72.45% using AIONER and MTRC for this step. Relaxing the evaluation criterion – to allow matching to either the annotated ChEBI or to a direct parent or child node in the ChEBI ontology – only slightly increased the overall performance of the end-to-end pipeline.

Normalization of protein mentions to UniProt ACs was also problematic. In addition to MTCR we attempted UniProtKB AC normalization using the recently developed gene normalization tool GNorm2 (https://github.com/ncbi/GNorm2) (53), which employs a straightforward Entrez ID to UniProtKB AC mapping table to link gene mentions to unique UniProtKB ACs. Performance of GNorm2 on EnzChemRED was relatively modest, with precision, recall, and $F_1$ score of only 39.66% / 11.43% / 17.74% respectively. GNorm2 utilizes Entrez IDs to map to UniProt ACs, and many enzymes in EnzChemRED have no Entrez ID. We also attempted mapping of gene and protein mentions using the UniProt API (https://www.uniprot.org/help/query-fields), retaining the first match as the UniProt AC for each protein name. NER+NEN performance for proteins remained marginal, with precision, recall, and $F_1$ score of 30.41%, 56.43%, and 39.52%, respectively. The $F_1$ score for ternary tuples was therefore significantly lower than that for binary pairs, at 20.74%.

### 4.6 Application of end-to-end pipeline to PubMed abstracts

We used the prototype end-to-end pipeline for enzyme function extraction described in the preceding section to extract candidate reaction pairs from 32 million PubMed abstracts (up to December 2023) (Table 7). We identified 680,426 relevant abstracts using our LitSuggest model, which together contained 158,837 distinct mentions of chemical pairs corresponding to one of the three valid relation types – "Conversion", "Indirect_conversion", or "Non_conversion" (Supplementary file 1).

**Table 7. Results of applying the end-to-end pipeline on PubMed abstracts.**

|  |  | Number | % of total |
|---|---|---|---|
| Abstracts | Total processed | 32,000,000 | 100 |
|  | Relevant according to LitSuggest | 680,426 | 2.13 |
|  | Containing predicted binary chemical- chemical relations (any type) | 64,077 | 0.20 |
| Binary chemical-chemical relations | All relation types | 176,084 | 100 |
| | Conversion | 158,837 | 90.21 |
| | Indirect_conversion | 16,889 | 9.59 |
| | Non_conversion | 358 | 0.20 |

We extracted all unique pairs of ChEBI IDs from relevant PubMed abstracts (Supplementary file 2) and compared them to unique pairs of ChEBI IDs extracted from Rhea reactions (Rhea release 130 of 8th November 2023) as described in Section 3.6 of Materials and methods. The results of this comparison are shown in Table 8.

**Table 8. Comparison of chemical conversion pairs from PubMed abstracts to Rhea.**

| | | In PubMed | In PubMed and in Rhea | % in Rhea |
|---|---|---:|---:|---:|
| Unique pairs of ChEBI IDs identified by our pipeline | All relation types | 37,715 | 3,152 | 8.36 |
| | Conversion | 30,661 | 2,721 | 8.87 |
| | Indirect_conversion | 6,986 | 428 | 6.13 |
| | Non_conversion | 68 | 3 | 4.41 |

Most of the chemical conversion pairs identified in PubMed abstracts – over 91% – are not found in Rhea and may correspond to novel chemical reactions that are potential candidates for curation in both Rhea and UniProtKB/Swiss-Prot. Figure 8 shows one approach to selecting potential high priority candidates among them, using a Tree MAP (TMAP) (62). This TMAP clusters those chemical conversion pairs from PubMed abstracts and from Rhea reactions where both members have an InChIKey according to the similarity of their differential reaction fingerprints (DRFP), which consider the differences in the circular substructures in the SMILES of each chemical pair (63). Clusters in the TMAP group chemical conversion pairs that have similar differences – chemical conversion pairs that, assuming that they are indeed pairs of reactants, would undergo similar changes during a reaction. To reduce the noise from the end-to-end pipeline we have applied an additional filter to the chemical conversion pairs from PubMed abstracts and Rhea reactions in the TMAP, based on their degree of atom conservation, as we would expect the main substrate-product pairs of curated reactions to share a high proportion of atoms on average, which would not be the case for erroneously identified chemical conversion pairs from PubMed abstracts. For Rhea reaction pairs, the mean atom conservation is around 79.76%, and we have applied this conservative threshold to filter pairs from PubMed abstracts and Rhea reactions during the construction of the TMAP shown in Figure 8. These types of filters can help curators to identify, and focus on, true novel chemical conversion pairs from the end-to-end pipeline.

Most of the chemical conversion pairs that are unique to PubMed abstracts form clusters in the TMAP with pairs from Rhea reactions, including several large clusters. This suggests that while the end-to-end pipeline may have identified many new chemical conversion pairs in PubMed, these generally undergo types of chemical conversion that have been previously curated in Rhea. The relatively low number of novel clusters from PubMed (all-blue clusters) that include no pairs from Rhea (red) suggests that coverage of published reaction chemistries in Rhea is already high, at least for those reactions that are described in abstracts and that involve chemical entities that can be mapped to ChEBI. Those all-blue clusters that are observed may represent novel reaction chemistries that are high priority targets for curation in UniProtKB/Swiss-Prot and Rhea, and curators are addressing these as a matter of priority. The 3,152 pairs of ChEBI IDs identified by our pipeline that map to Rhea covered only 3,800 reactions of the total of 16,112 Rhea reactions at the time of analysis (Rhea release 130). The Rhea reactions that were not identified by our pipeline may have been missed due to low recall of our methods, or may only be described in full text and/or figures, which provide more information for NLP pipelines (66).

## 5. Discussion

This work describes EnzChemRED, a high-quality expert curated corpus of PubMed abstracts designed to support the development and benchmarking of NLP methods to extract knowledge of enzyme functions from scientific literature. We demonstrate the benefits of using EnzChemRED by fine-tuning pre-trained language models for NER and RE and combining these models in a prototype end-to-end pipeline that was used to build a draft map of enzyme functions from PubMed abstracts. While EnzChemRED was originally conceived to support the curation of enzyme functions in UniProtKB/Swiss-Prot using the reaction knowledgebase Rhea and the chemical ontology ChEBI, we hope that EnzChemRED can serve as a useful training and benchmarking set for NLP method developers and for other knowledgebases and resources.

### 5.1 Limitations

EnzChemRED has two main limitations. First, EnzChemRED was developed using PubMed abstracts rather than full text. Full text contains more information than abstracts (66), but may not be freely available for many articles. EnzChemRED is also fairly small compared to some RE datasets (Table 1), consisting of only 1,210 PubMed abstracts, but our work seems to indicate that EnzChemRED is already comprehensive enough to support the fine-tuning of language models that perform well at both NER and RE.

Tests of our end-to-end pipeline on EnzChemRED revealed limitations of the methods used, with NEN of gene/protein mentions to UniProtKB ACs using MTCR and GNorm2 proving to be a major challenge. Many of the proteins in EnzChemRED lack the Entrez IDs used in GNorm2, and we plan to expand the gene vocabulary of this tool using UniProt and other data sources. New corpora for species recognition may further assist gene and protein name disambiguation (67). Normalization of chemical mentions to ChEBI IDs was also challenging due to the high complexity of the ChEBI ontology and the propensity of authors to use ambiguous chemical names that may be applicable to several isomers, including stereoisomers, as well as to related compounds or classes of compound, within the ChEBI ontology. For example, authors generally refer in abstracts to L-amino acids without specifying stereochemistry, so "L-serine" would generally be referred to as "serine", which our MTCR approach mapped to CHEBI:35243, a serine zwitterion of undefined stereochemistry. In contrast in the EnzChemRED corpus mentions of

"serine" were curated to CHEBI:33384, the "L-serine zwitterion", when information in the full text and other sources allowed. We found that allowing fuzzy matching to parents or children in the ChEBI ontology during NEN improved performance by a small margin (**Table 6**). We also expect that performance of NEN methods for chemicals may be lower for PubMed abstracts generally than for those abstracts that form part of EnzChemRED. The latter were drawn from publications used to curate UniProtKB/Swiss-Prot entries with Rhea reactions, meaning that the chemical mentions in EnzChemRED abstracts would have been targeted for curation in ChEBI; we would therefore expect a lower rate of false negatives for NEN on EnzChemRED abstracts than for NEN on PubMed abstracts. One route to reducing FNs in chemical NEN may be to use larger chemical dictionaries such as PubChem, which we intend to explore. The best solution would be for authors to use unambiguous machine readable chemical identifiers such as SMILES or InChIs/InChIKeys, and some journals now recommend their use in publications (68). Methods to extract reaction data from figures (69,70) may also provide a useful complement to the text-based methods tested here.

## 5.2 Future Works

We will continue to develop the EnzChemRED corpus in response to feedback from users and to work to improve the methods for protein and chemical NER and NEN used in our prototype end-to-end pipeline. One area we are exploring is to extend the curation of EnzChemRED to include mappings of sentences and larger passages to other vocabularies, including Gene Ontology terms and the hierarchical enzyme classification of the IUBMB (EC numbers), to test embeddings and other approaches for semantic annotation of text. UniProt and Rhea curators are now using the output from the end-to-end pipeline to guide the curation of new enzymes and reactions in UniProtKB/Swiss-Prot and Rhea, including to link disconnected reactions in Rhea and UniProtKB/Swiss-Prot, analogous to gap-filling for genome scale metabolic models (71). We are also testing the ability of large language models (LLMs) such as GPT-3.5 and GPT-4 to perform the RE task described here, as well as to extract knowledge of enzymes and their reactions directly using other kinds of simple prompt, with zero- and few-shot learning and fine-tuning. While smaller domain-specific models such as PubMedBERT and BioREx have to date proven to be at least as effective as much larger commercially available LLMs for most common NLP tasks tested (72,73), the possibility to further fine-tune LLMs using curated datasets like EnzChemRED may offer a route to creating high performing RE systems using smaller training datasets, and to the augmentation of training datasets by rephrasing.

**Key Points**

- We present EnzChemRED, a new enzyme function relation extraction dataset with annotation of gene/protein and chemical mentions and the binary and ternary relations that define chemical conversions that characterize enzymatic reactions and the enzymes that catalyze them, at sentence and mention level.
- We demonstrate the utility of EnzChemRED to fine tune and evaluate methods for NER and RE for enzyme functions and combine these methods into a prototype end-to-end pipeline for enzyme function extraction from text. While promising, the results show that there is much room for improvement, particularly for NEN of proteins and chemicals.

- We showcase a draft map of enzyme functions that can be used to guide curation of Rhea and UniProtKB/Swiss-Prot to unexplored areas of enzyme function space in literature.

## Data availability

The EnzChemRED corpus is available for download in BioC format (74) at the Rhea ftp site: https://ftp.expasy.org/databases/rhea/nlp/.

The PubMedBERT and BioREx models are available at https://github.com/ncbi/biored and https://github.com/ncbi/BioREx.

The LitSuggest model described in this work is available at https://www.ncbi.nlm.nih.gov/research/litsuggest/project/5fa57e75bf71b3730469a83b.

## Supplementary information

**Supplementary file 1** (https://ftp.expasy.org/databases/rhea/nlp/BioREx_EnzChemRED_PubMed.tsv) provides chemical conversion pairs (binary pairs) identified in PubMed using our prototype end-to-end pipeline, including the PMID and sentence from which they derive, and their BioREx score.

**Supplementary file 2** (https://ftp.expasy.org/databases/rhea/nlp/BioREx_EnzChemRED_PubMed_normalized.tsv) provides unique chemical conversion (binary pairs) identified in PubMed using our prototype end-to-end pipeline, with key characteristics such as PubMed count, maximum BioREx score, atom conservation, and the minimum DRFP distance to a ChEBI pair in Rhea.

## Acknowledgements

We would like to thank the Cheminformatics and Metabolism Team of EMBL-EBI for their work in maintaining and developing ChEBI, without which Rhea would not be possible, particularly Adnan Malik for expert curation. We are indebted to Blanca Cabrera Gil at Swiss-Prot for stimulating discussions and advice on machine learning in chemistry. We gratefully acknowledge the software contributions of Chemaxon (https://www.chemaxon.com/products/marvin/).

## Funding

Expert curation and evaluation of EnzChemRED at Swiss-Prot were supported by the Swiss Federal Government through the State Secretariat for Education, Research and Innovation (SERI) and the National Human Genome Research Institute (NHGRI), Office of Director [OD/DPCPSI/ODSS], National Institute of Allergy and Infectious Diseases (NIAID), National Institute on Aging (NIA), National Institute of General Medical Sciences (NIGMS), National Institute of Diabetes and Digestive and Kidney Diseases (NIDDK), National Eye Institute (NEI), National Cancer Institute (NCI), National Heart, Lung, and Blood

Institute (NHLBI) of the National Institutes of Health [U24HG007822], and by the European Union's Horizon Europe Framework Programme (grant number 101080997), supported in Switzerland through the State Secretariat for Education, Research and Innovation (SERI).

This research was also supported by the NIH Intramural Research Program, National Library of Medicine and by the Fundamental Research Funds for the Central Universities [DUT23RC(3)014 to L.L.].

Funding for open access charge: NIH.

```
PREFIX rh: <http://rdf.rhea-db.org/>
   PREFIX rdfs: <http://www.w3.org/2000/01/rdf-schema#>
   PREFIX rdf: <http://www.w3.org/1999/02/22-rdf-syntax-ns#>
   PREFIX ECO: <http://purl.obolibrary.org/obo/ECO_>
   PREFIX up: <http://purl.uniprot.org/core/>
   SELECT ?rhea ?catalyzedReaction ?source
   WHERE {
      {
         SERVICE <https://sparql.rhea-db.org/sparql> {
            SELECT DISTINCT ?rhea
            WHERE {
               ?rhea rdfs:subClassOf rh:Reaction .
               ?rhea rh:side/rh:contains/rh:compound ?compound .
               ?compound rh:chebi ?chebi ;
               rdfs:subClassOf rh:SmallMolecule .
            }
         }
      }
      ?catalyzedReaction up:catalyzedReaction ?rhea .
      ?reif rdf:object ?catalyzedReaction ;
      up:attribution ?attr .
      ?attr up:evidence ECO:0000269 ;
      up:source ?source .
      ?source a up:Citation .
   }
```

**Figure 1. SPARQL query used to identify papers for abstract curation in EnzChemRED.**

| Chemical | Protein |
|---|---|
| "NADP(+)," "succinate," "cis-3-hydroxy-l-proline," and "3- and 4-hydroxyprolines" | "Maf," "PLR-Lp1," "A0NXQ8," "E. C. 2.8.3.5," "10-formyltetrahydrofolate dehydrogenase," and "erythromycin O-methyltransferase" |

| Domain* | MutantEnzyme* | Coreference* |
|---|---|---|
| "C domain," "epimerization (E) domain," "N-terminal MIT-like domain," and "epimerization (E) domain" | "C160A mutant," "mmcR-deletion," and "K240M YPK" | "they," "it," "the enzyme," "the hydrolase," and "the recombinant enzyme" |

**Figure 2. Entity curation in EnzChemRED.** We curated all Chemical and Protein mentions, but not of Domain, MutantEnzyme and Coreference (denoted by '*'), for which curation focused on those mentions that participate in conversions. The latter are not included in our evaluations.

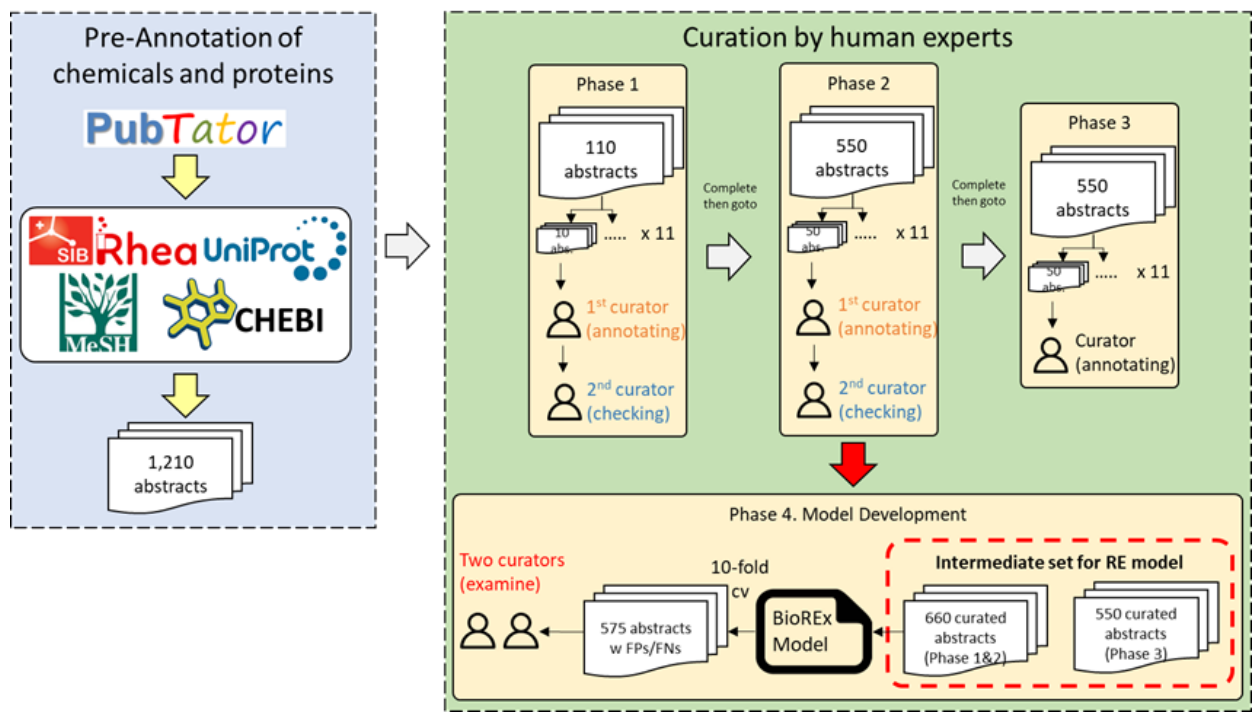

**Figure 3 – Curation workflow for the EnzChemRED corpus.** A total of 1,210 abstracts were curated by 11 experts in three phases. These 1,210 abstracts were then used to train an interim BioREx model, which was then run on all EnzChemRED abstracts; abstracts with putative FP and FN predictions by the model were then analyzed again in phase 4, and, where necessary, re-curated.

**Figure 4. Annotation of EnzChemRED abstracts using TeamTat.** In the abstract shown (from reference (75)) the mentions of "Cysteine dioxygenase" and "CDO" refer to the enzyme (UniProt:P60334) that is responsible for the conversion of L-cysteine (CHEBI:35235) and cysteine sulfinic acid (CHEBI:61085). The inset shows details of the curated relation within the TeamTat tool, including the type of relation ("Conversion"), the text spans that define the participants in that relation, and their offsets, the unique identifiers from UniProtKB and ChEBI that were used to tag those text spans, and the assignment of the role "Converter" to the text spans of the protein mentions.

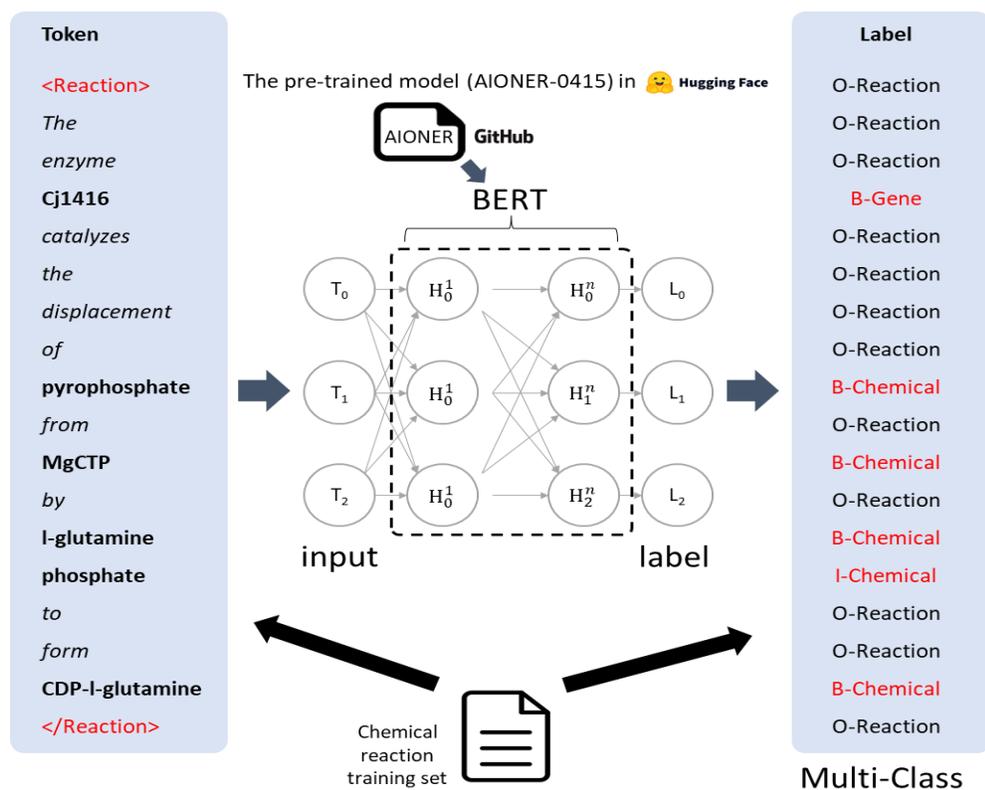

Figure 5. Overview of the fine-tuning process for AIONER on EnzChemRED.

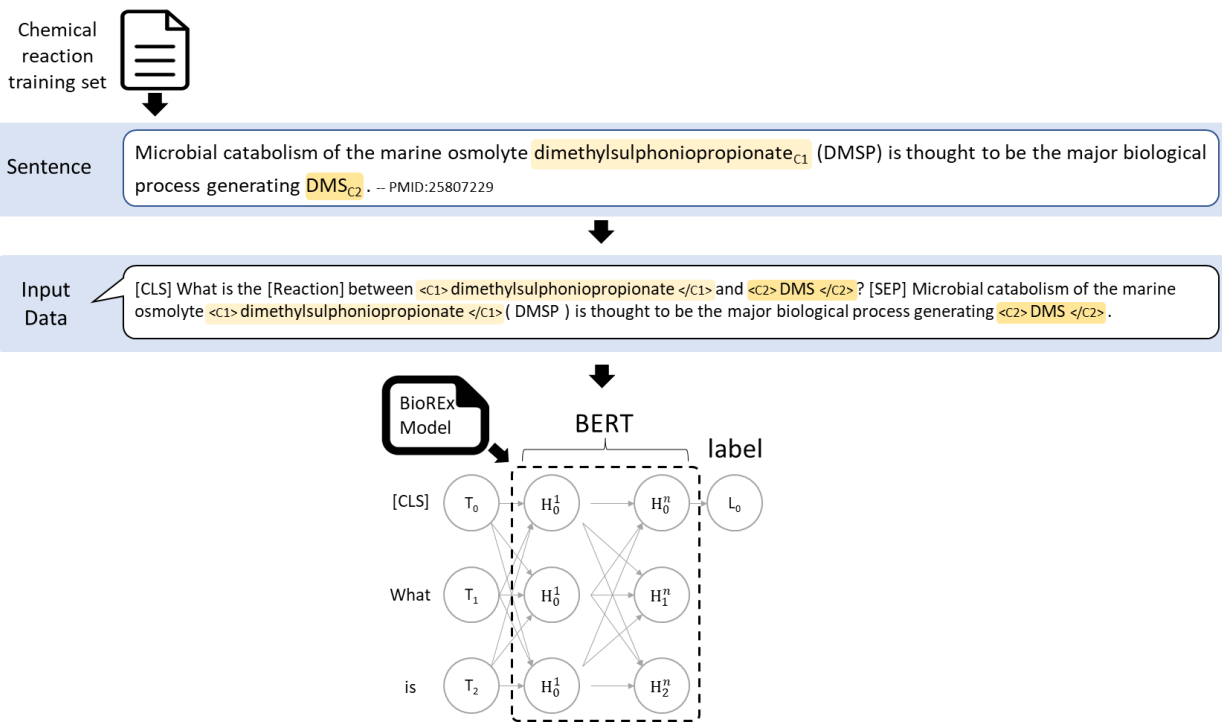

Figure 6. An illustration of the specific input representation for the EnzChemRED dataset and the fine-tuning process of the BioREx model.

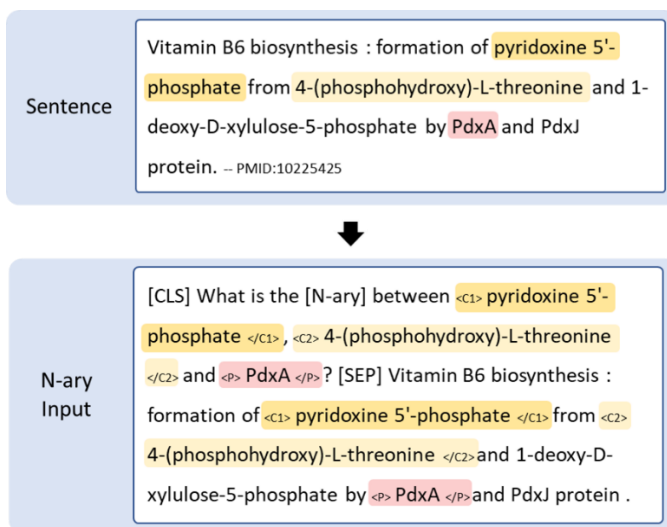

**Figure 7.** An example of the input representation for a ternary tuple in EnzChemRED.

A

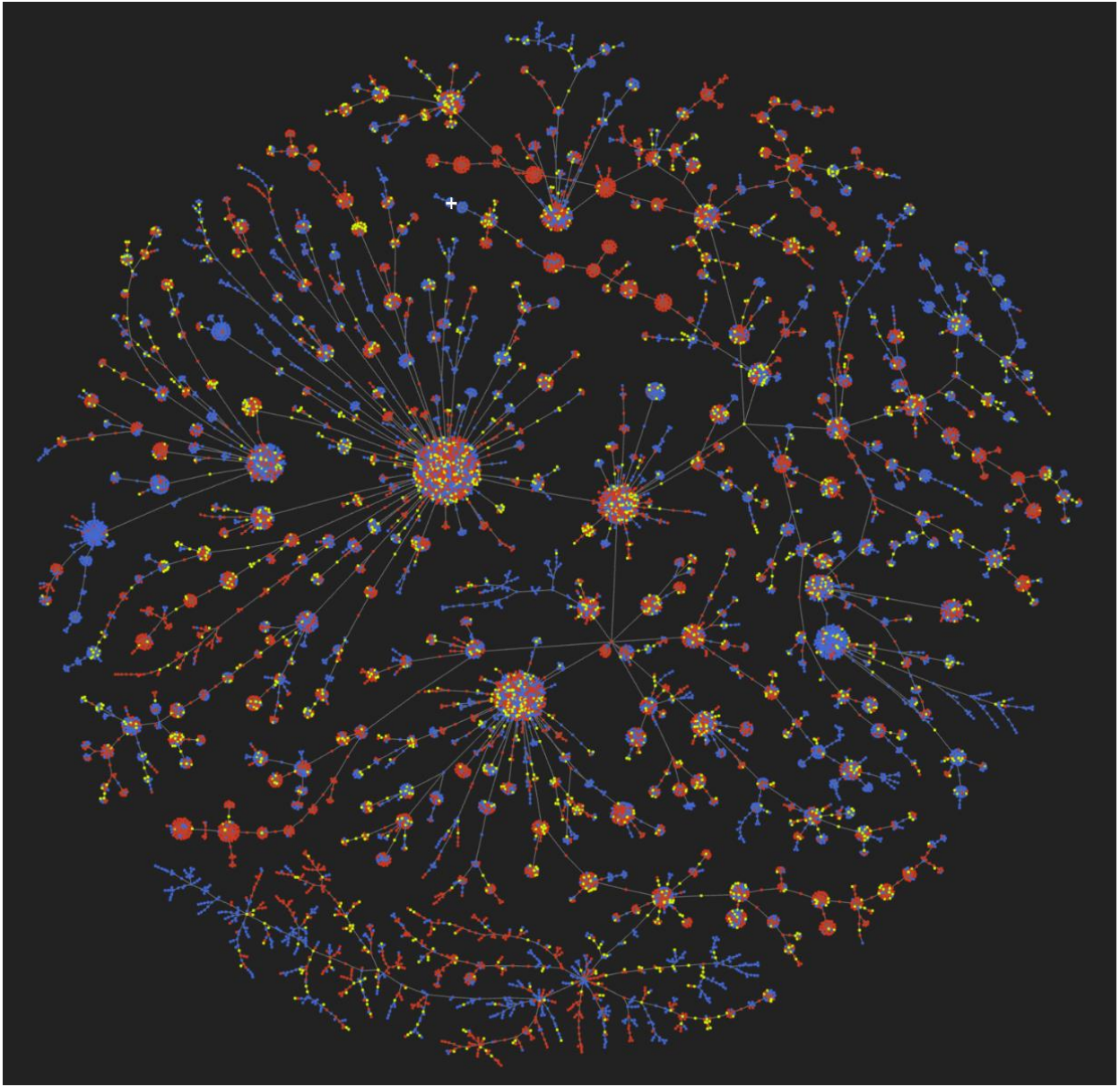

B

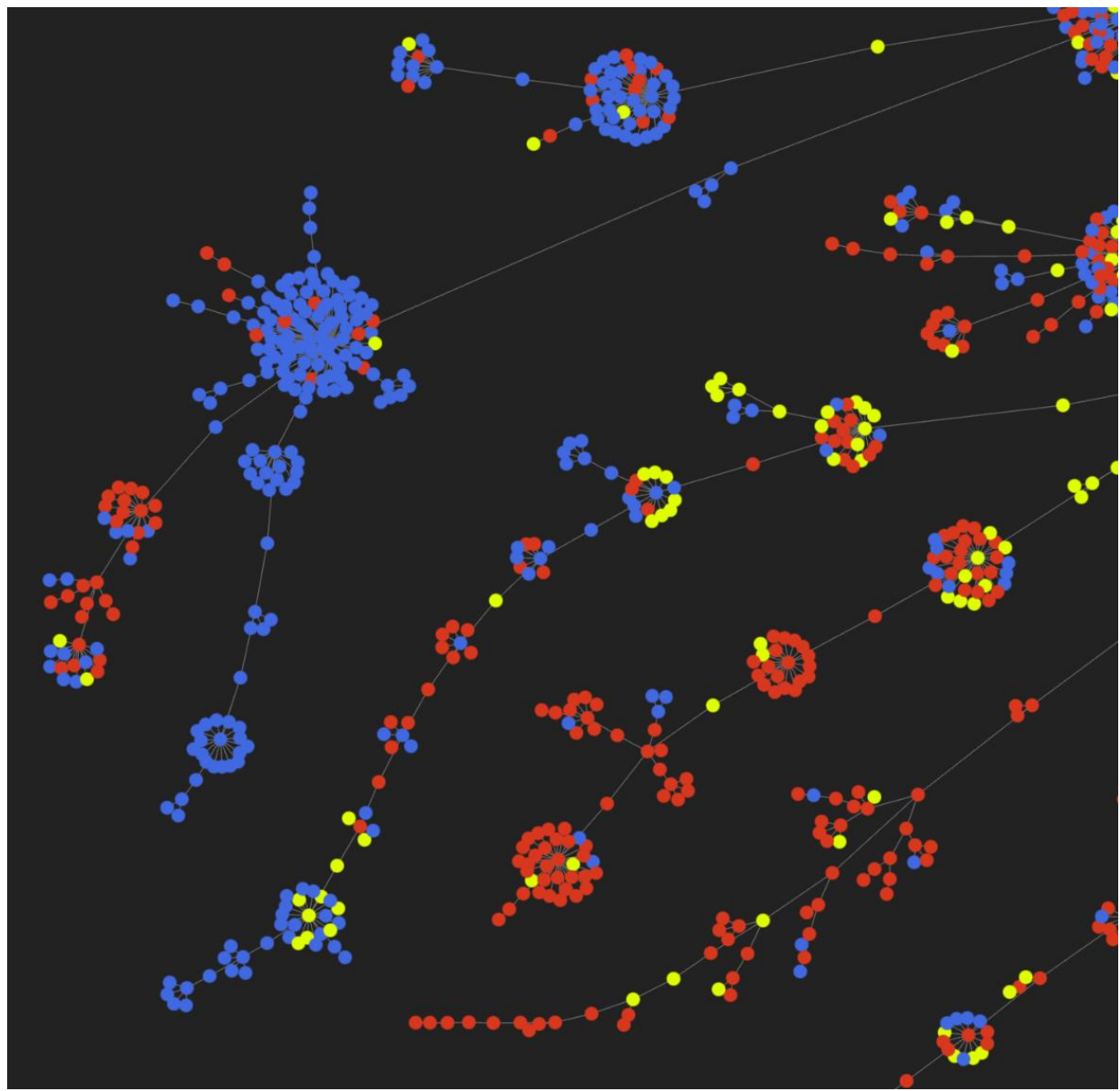

**Figure 8. TMAP of chemical conversion pairs extracted from PubMed abstracts and from Rhea.** Each point in the TMAP corresponds to a chemical conversion pair, which are clustered according to their Differential Reaction Fingerprint (DRFP). Blue, chemical conversion pairs from PubMed abstracts only; red, chemical conversion pairs from Rhea reactions only; yellow, chemical conversion pairs common to both PubMed abstracts and Rhea (as determined by ChEBI ID matching). A) View of complete TMAP. B) Zoom on the TMAP reveals further details of clustering of chemical pairs. Novel chemical pairs from PubMed abstracts, with chemical differences that differ from those curated in Rhea, appear as blue nodes that do not cluster with red, and may be one priority for curation at Rhea and UniProt.